\documentclass[letterpaper, 10 pt, journal, twoside]{ieeetran}  % Comment this line out if you need a4paper

\IEEEoverridecommandlockouts                              % This command is only needed if 
\usepackage{graphics} % for pdf, bitmapped graphics files
\usepackage{epsfig} % for postscript graphics files
\usepackage{mathptmx} % assumes new font selection scheme installed
\usepackage{times} % assumes new font selection scheme installed
\usepackage{amsmath} % assumes amsmath package installed
\usepackage{amssymb}  % assumes amsmath package installed
\usepackage{algorithm}
\usepackage{algpseudocode}
\usepackage{multirow}
\usepackage{booktabs}
\usepackage{caption}
\usepackage{subcaption}
\usepackage{enumitem}
\usepackage[usenames,dvipsnames]{color}
\usepackage[noadjust]{cite}

\title{DeRi-Bot: Learning to Collaboratively Manipulate \\Rigid Objects via Deformable Objects}

\author{Zixing Wang and Ahmed H. Qureshi
\thanks{Manuscript received: Apr, 19, Year; Revised July, 18, 2023; Accepted Aug, 14, 2023.}
\thanks{This paper was recommended for publication by Editor Tetsuya Ogata upon evaluation of the Associate Editor and Reviewers' comments.} %Use only for final RAL version
\thanks{Zixing Wang and Ahmed H. Qureshi are with the Department of Computer Science, Purdue University, West Lafayette, IN 47907, USA
        {\tt\footnotesize \{wang5389, ahqureshi\}@purdue.edu}}%
\thanks{Digital Object Identifier (DOI): see top of this page.}}%Use only for final RAL version

\begin{document}
\maketitle

%%%%%%%%%%%%%%%%%%%%%%%%%%%%%%%%%%%%%%%%%%%%%%%%%%%%%%%%%%%%%%%%%%%%%%%%%%%%%%%%
\begin{abstract}
Recent research efforts have yielded significant advancements in manipulating objects under homogeneous settings where the robot is required to either manipulate rigid or deformable (soft) objects. However, the manipulation under heterogeneous setups that involve both rigid and one-dimensional (1D) deformable objects remains an unexplored area of research. Such setups are common in various scenarios that involve the transportation of heavy objects via ropes, e.g., on factory floors, at disaster sites, and in forestry. To address this challenge, we introduce DeRi-Bot, the first framework that enables the collaborative manipulation of rigid objects with deformable objects. Our framework comprises an Action Prediction Network (APN) and a Configuration Prediction Network (CPN) to model the complex pattern and stochasticity of soft-rigid body systems. We demonstrate the effectiveness of DeRi-Bot in moving rigid objects to a target position with ropes connected to robotic arms. Furthermore, DeRi-Bot is a distributive method that can accommodate an arbitrary number of robots or human partners without reconfiguration or retraining. We evaluate our framework in both simulated and real-world environments and show that it achieves promising results with strong generalization across different types of objects and multi-agent settings, including human-robot collaboration.
\end{abstract}


%%%%%%%%%%%%%%%%%%%%%%%%%%%%%%%%%%%%%%%%%%%%%%%%%%%%%%%%%%%%%%%%%%%%%%%%%%%%%%%%
\section{Introduction}
\label{sec:intro}

\begin{figure}[t]
         \includegraphics[width=0.48\textwidth]{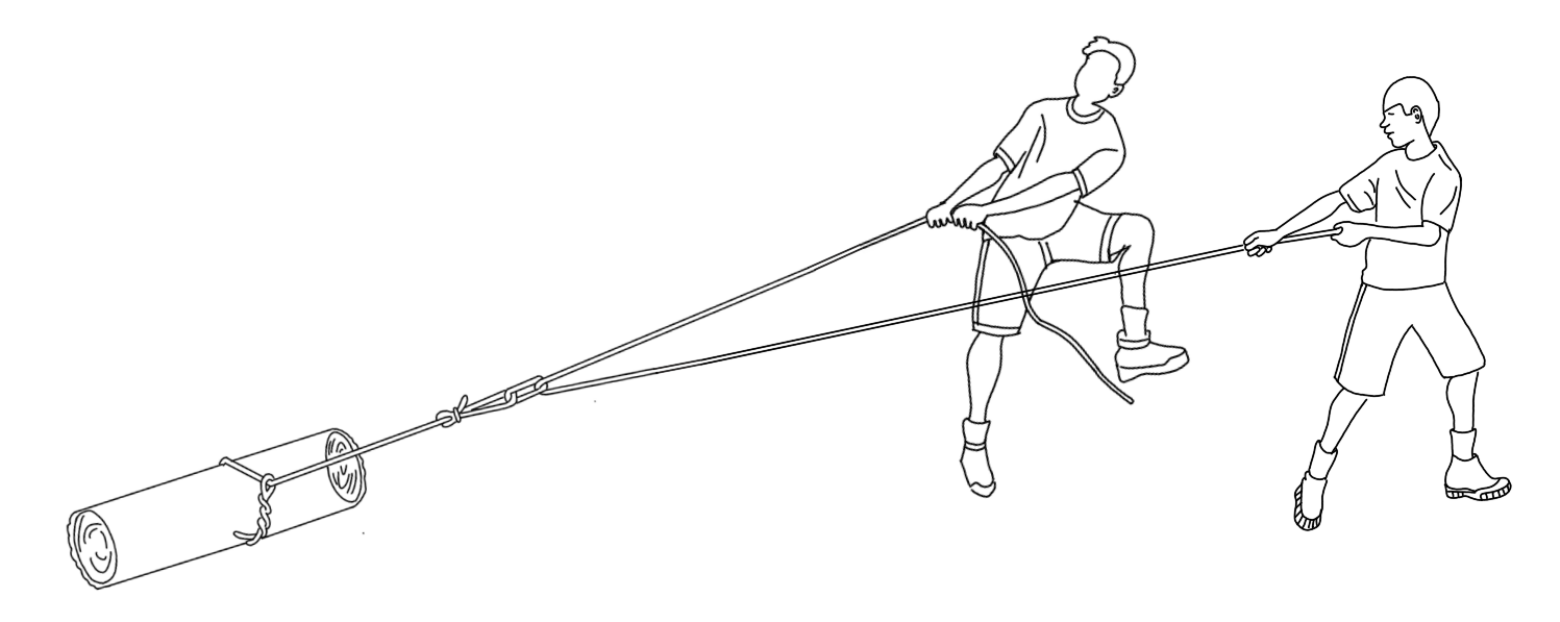}
         \caption{A real-world scenario of moving a rigid object (tree trunk) via soft objects (ropes).}
         \label{subfig:title_up}
         \vspace{-4mm}
\end{figure}

\IEEEPARstart{T}{he} manipulation of heterogeneous soft-rigid systems, which involve connected deformable and rigid bodies, is crucial for various tasks. For instance, in forestry, ropes are often used to transport logs or large tree branches from one place to another, as illustrated in Fig.~\ref{subfig:title_up}. Similarly, in disaster response scenarios, first responders may need to use ropes to move debris or heavy objects to search for victims or clear the path for rescue vehicles. Therefore, designing robots that can manipulate connected rigid and deformable objects in such scenarios can significantly enhance the efficiency and safety of operations.

Despite various applications of manipulating soft-rigid object systems, the existing works have primarily explored homogeneous settings where robots manipulate either rigid or deformable objects~\cite{billard2019trends, sanchez2018}. For instance, recent research has demonstrated impressive capabilities in rigid body manipulation, such as novel objects grasping~\cite{keshari2022cograsp} and tool-based manipulation~\cite{9196971, xu2021deep, xie2019improvisation}. Some studies have also explored deformable object manipulation, with~\cite{ha2021flingbot} using a manipulator to effectively unfold clothing and~\cite{zhang2021robots} learning to wave ropes to vault and knock objects.
%~\cite{xu2022dextairity} utilizing air currents for manipulating deformable bodies. 

\begin{figure}[t]
  \centering
         \includegraphics[width=0.48\textwidth]{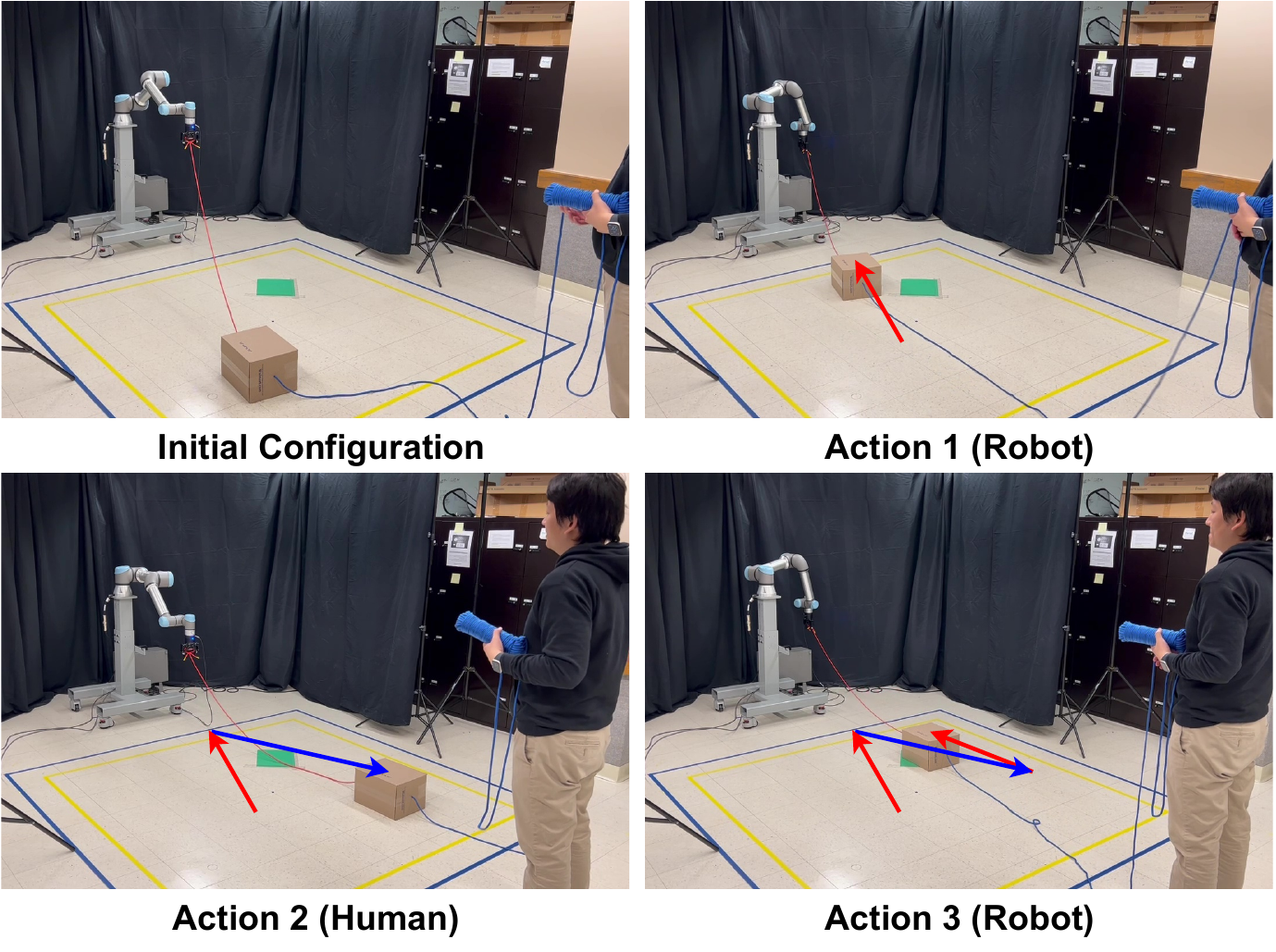}         
         \caption{A demonstration of DeRi-Bot collaborating with a human in the real world in moving the rigid brown box to its target position given by a green marker.}
         \label{subfig:title_down}
         \vspace{-8mm}
\end{figure}

%\begin{figure}[t]
%  \centering
%  \vspace{6pt}
%    \begin{subfigure}[b]{0.48\textwidth}
%         \centering
%         \includegraphics[width=\textwidth]{imgs/hh_pulling.png}
%         \caption{A real-world scenario of moving a rigid object (tree trunk) via soft objects (ropes).}
%         \label{subfig:title_up}
%    \end{subfigure}
 %   \par\bigskip
%    \begin{subfigure}[b]{0.48\textwidth}
%         \centering
%         \includegraphics[width=\textwidth]{imgs/title.pdf}         
%         \caption{A demonstration of DeRi-Bot collaborating with a human in the real world in moving the rigid brown box to its target position given by green marker.}
%         \label{subfig:title_down}
%    \end{subfigure}
%    \caption{DeRi-Bot is inspired by real-world applications with the objective of enabling collaborative soft-rigid body manipulation for robots}
%\end{figure}

In this paper, we propose the first \textbf{De}formable-\textbf{Ri}gid Ro\textbf{bot} (DeRi-Bot) framework that enables a rigid body manipulation with 1D deformable objects, such as ropes. Specifically, moving a rigid object to designated positions by pulling ropes, as shown in Fig.~\ref{subfig:title_down}. However, designing such framework needs to overcome the following challenges presented by soft-rigid systems. Firstly, deformable bodies possess significantly more degrees of freedom than rigid bodies, which makes explicit and numerical modeling of their movement considerably more complex and expensive. Secondly, the complexity of soft-rigid systems prohibits direct scaling of the connected rigid bodies' size and shape to the system control signal, leading to generalization problems. Additionally, the inherent deformability makes soft bodies' movement behavior highly unpredictable and introduces multi-solution problems, which can be roughly viewed as a stochastic system. Finally, when dragging rigid objects via ropes, multiple agents, such as humans or robots, often collaborate to move the object, as illustrated in Figs.~\ref{subfig:title_up}-\ref{subfig:title_down}. Therefore, the framework must be able to operate in a team with any number of agents to achieve efficient manipulation of soft-rigid systems. 

These challenges necessitate the following components for the design of DeRi-Bot. (1) \textbf{Action Prediction Network} (APN). It predicts the required robot action and implicitly models the latent pattern between the soft-rigid body system and robot control commands for manipulating arbitrary-sized rigid bodies via ropes, thus tackling the first two challenges mentioned above. %which allows us to get rid of the complex physics modeling work. 
(2) \textbf{Configuration Prediction Network} (CPN). Because of the high stochasticity of soft bodies (third challenge), we intend to explore solutions in a wider range. Thus, we improvise by sampling commands from a Gaussian distribution centered around the output of APN. Given the sampled and APN generated action commands and the current environment state, CPN predicts the corresponding outcome configuration of the soft-rigid body, allowing the selection of the best action for execution that leads to the given goal state. (3) Lastly, to overcome the problem introduced by the need for a collaborative system, our framework decouples the synchronous task into an asynchronous independent task, i.e., only one agent acts at a time, and we leverage our CPN to select the agent whose action leads the rigid object closer to its target. This allows multiple agents to collaborate asynchronously to achieve the target. We integrate all the above-mentioned components into a unified framework, DeRi-Bot, and evaluate it in simulated and real-world environments. The results indicate that the proposed framework achieves all our objectives and generalizes to various system setups, including multiagent and human-robot collaboration settings, without needing any reconfiguration or retraining of the proposed framework. 

To summarize, the main contributions of this work are as follows:
\begin{itemize}[leftmargin=*]
    \item We propose DeRi-Bot, the first framework that enables collaborative manipulation of rigid bodies using 1D deformable bodies.
    \item We develop APN to implicitly model the underlying pattern governing the relationship between soft-rigid body systems and robot action commands.
    \item We design CPN to obtain visual foresight and overcome the challenge of rope stochasticity. This results in a robust framework capable of predicting the soft-rigid object configurations from different robot actions and selecting the best action for execution to reach the target position.
    \item A strategy to train the DeRi-Bot system asynchronously such that the proposed solution can scale to an arbitrary number of agents manipulating the rigid object via ropes.
\end{itemize}

\begin{figure*}[t]
  \centering
  \vspace{10pt}
  \includegraphics[width=\textwidth]{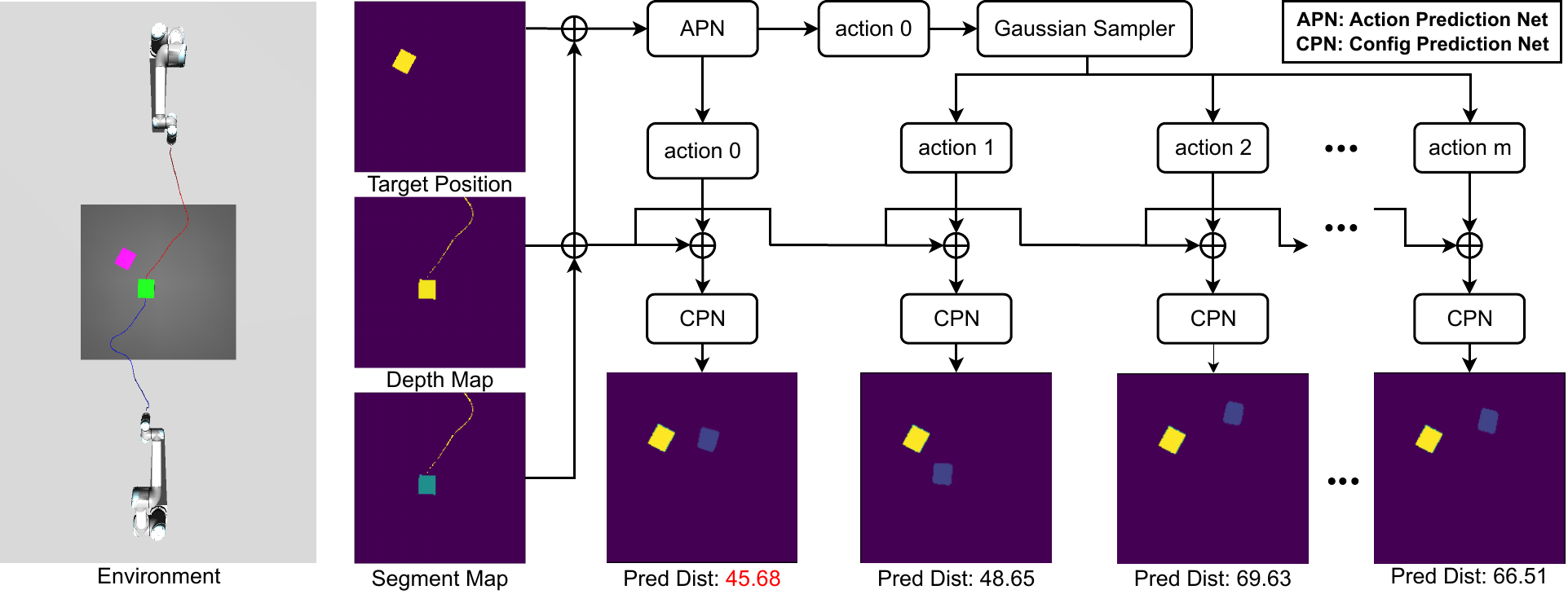}
  \caption{An iteration of DeRi-Bot workflow. The APN model takes as input the state observation to predict an action. The Sampler generates more actions around the output of APN. The CPN model predicts the corresponding next states of the rigid object of all the actions. The framework picks the best action that yields the minimum distance between the object and the target position. This figure shows such a process for the robot on the top. However, by design, all the robots' proposed actions will be compared together for the selection of the robot and its best action. Furthermore, note that we add the target in yellow to the visualization of CPN outputs to illustrate the effect of different actions.}
  \label{fig:workflow}
  \vspace{-5mm}
\end{figure*}

\section{Related Works}
Perhaps~\cite{donald2000distributed} is the closest work to ours. It proposes a manually designed model-based multi-agent algorithm for packing and moving multiple objects via ropes. Although this method solves the task of transporting objects via ropes, this leverages manually designed action primitives. In contrast, our approach is a data-driven method and can generalize to a wide range of scenarios that otherwise are difficult to envision. Since DeRi-Bot is the first data-driven method focusing on heterogeneous soft-rigid body manipulation, we present the existing works solving the homogeneous task of manipulating rigid or soft objects and draw their relevance and use cases to our proposed framework.
\subsection{Rigid Body Manipulation}
Plenty of work has been done in rigid body manipulation~\cite{9196971, xu2021deep, xie2019improvisation, 9832465, agrawal2016learning, Stoytchev2005BehaviorGroundedRO, toussaint2018differentiable}, but none of these approaches consider the task of using soft objects such as ropes for manipulating rigid bodies. The use of rigid objects to manipulate other rigid objects has also been widely explored. For instance, ~\cite{toussaint2018differentiable} proposes a Task and Motion Planning (TAMP) technique for using rigid objects to interact with other rigid objects, such as using a hook to drag another rigid object closer to the robot. Another method introduces a Deep Affordance Foresight~\cite{xu2021deep} that predicts the long-term effects of various actions, including tool use for moving other rigid objects. Likewise,~\cite{agrawal2016learning} performs the poking action with a stick to move rigid objects and develop a visual foresight of the environment. Moreover, manipulation with non-trivial tools has also been investigated. For example,~\cite{kawaharazuka2020tool} uses a neural network to optimize the shape and motion of a rigid planar tool represented by an image and tests it on 2D objects' transportation tasks. Similar to these approaches, we also build a visual foresight module, but our approach considers a 1D deformable object, i.e., rope, to manipulate a rigid object, which introduces the complexity of dealing with soft-object physics and its impact on the rigid objects' motion. 

\subsection{Soft Body Manipulation}

% We point readers a survey~\cite{sanchez2018} that gives a comprehensive review on recent advancement in this field. Another comprehensive work is SoftGym~\cite{corl2020softgym}, which provides various deformable body manipulation tasks and benchmark with recent baselines

Deformable object manipulation is a challenging problem due to its high degree of freedom. Plenty of work has been introduced ranging from imitation learning to reinforcement learning to model deformable object dynamics~\cite{Lee2014UnifyingSR,nair2017combining,kawaharazuka2019dynamic}. For example,~\cite{Lee2014UnifyingSR} optimizes learning from the demonstration for generalization to the novel deformable object manipulation task.~\cite{nair2017combining} uses self-supervised and imitation learning to manipulate the rope to place it in the given target configuration. In addition,~\cite{kawaharazuka2019dynamic} proposes a framework to generate an optimized time-series joint torque command to align the rope with the target 2D configuration. Other than that, demonstration-free methods such as~\cite{wu2020} leverage model-free reinforcement learning to manipulate 1 to 2-dimensional deformable objects, and~\cite{zhang2021robots} use a self-supervised framework to plan for high-speed rope manipulation. The application of modeling deformable object physics has been demonstrated in various tasks such as knot-tying with a rope~\cite{Hopcroft1991ACS, 7487548, Morita2003} and rope untangling~\cite{6696448, Grannen2020UntanglingDK, she2021cable}. Although the methods in modeling deformable object dynamics may be helpful in our approach to predict desired rope configurations for moving the attached rigid object, we follow a model-free approach that directly trains a robot action policy and implicitly learns the deformable object physics along with their relation to rigid-object and robot action dynamics. Perhaps a more relevant approach to our method is the iterative residual method~\cite{chi2022irp}. Similar to CPN, it leverages the Delta Dynamics Network to predict the outcome rope tip trajectory after adjusting the previous whipping action. By iteratively applying the best adjustments to the last action, the rope tip eventually reaches the target position. Similar to their approach, our method also forecasts object states given different actions, but we evaluate actions supported by neural networks rather than random residual actions and consider a heterogeneous setup that combines both rigid and soft objects.

\section{Methods}
In this section, we present our methodology, DeRi-Bot, to achieve our objectives of designing a system that can work with an arbitrary number of agents to collaborate in moving rigid objects to their goals via rope manipulation. To account for multiagent settings, we formalize a decoupling strategy that allows our framework trained on a single agent to generalize to multiagent scenarios asynchronously. Specifically, given the environment's current state, only one of the agents can execute its proposed action at a given time step. The agent for executing the action is selected heuristically or by a human's decision. The process is repeated iteratively, modifying the object's position to minimize the distance to the given target position. Due to decoupling strategy, DeRi-Bot possesses the following advantages:
\begin{itemize}[leftmargin=*]
    \item It can accommodate any number of robots of the same type without retraining or reconfiguration.
    \item It has the flexibility to work with human actors whose actions are hard to predict and estimate.
\end{itemize}
%Figs.\ref{subfig:title_down} and~\ref{fig:dual-exp} show, our task requires participants to collaboratively manipulate their ropes to move the object to the target position. However, because of the complex properties of soft bodies and the hard-to-model human behaviors, the typical synchronous commanding primitive will bring significant unnecessary complexity, despite the potential time and step savings.
%\subsubsection{Decoupling}
%To overcome these problems, we propose to decouple the synchronous problem into a series of independent sub-tasks. Specifically, given the environment's current state, only one participant executes its proposed action, where the acting participant at a given instance is determined by a scoring function or a human's decision. The process is repeated, iteratively modifying the object's position, to minimize the distance between the object and target position. DeRi-Bot possesses the following advantages with our decoupling strategy:
%\begin{itemize}
%    \item The trained networks can accommodate a various number of robots of the same type without retraining or reconfiguration.
%    \item The framework has the flexibility to work with human actors, whose actions are hard to predict and estimate.
%\end{itemize}
In the remainder of this section, we describe our problem statement and DeRi-Bot's components including it's neural models, workflow, and training strategy.  
\subsection{Problem Statement}
\label{subsec:pds}
  %Figs.\ref{subfig:title_down} and~\ref{fig:dual-exp} show, our task requires participants to collaboratively manipulate their ropes to move the object to the target position. However, because of the complex properties of soft bodies and the hard-to-model human behaviors, the typical synchronous commanding primitive will bring significant unnecessary complexity, despite the potential time and step savings.
%\subsubsection{Decoupling}
%To overcome these problems, we propose to decouple the synchronous problem into a series of independent sub-tasks. Specifically, given the environment's current state, only one participant executes its proposed action, where the acting participant at a given instance is determined by a scoring function or a human's decision. The process is repeated, iteratively modifying the object's position, to minimize the distance between the object and target position. DeRi-Bot possesses the following advantages with our decoupling strategy:
%\begin{itemize}
%    \item The trained networks can accommodate a various number of robots of the same type without retraining or reconfiguration.
%    \item The framework has the flexibility to work with human actors, whose actions are hard to predict and estimate.
%\end{itemize}

We formulate the task with our proposed asynchronous action primitive as follows. An instance of the experimental environment contains one object, one target indicator, and an arbitrary number of agents connected to the object via ropes (one rope per agent). We use the following notation to represent their states. $d \in D$ denotes the orthographic depth map of the task environment, where $D \subset \mathbb{R}^{n \times n}$ and $n$ indicates the map dimension. We regularize $D$ by setting the ground plane as 0 meters. Similarly, $s \in S$ denotes the segmentation map of the environment, where $S \subset \{0, 1, 2\}^{n \times n}$, in which 0, 1, and 2 represent the ground plane, the rope, and the object. A variant of $S$ that is without the encoding of rope is represented as $\hat{s} \in S$. We use a binary map $b \in B$ to encode the target position, where $B \subset \{0, 1\}^{n \times n}$, and pixels belonging to the target are indicated by 1's. Both $S$ and $B$ are defined in the same domain as $D$. Let each agent be denoted by $c \in C$. Each agent's desired end-effector position is indicated by $a \in A$, where $A\subset \mathbb{R}^3$ represents the 3D workspace. %The mappings between $C$ and $A$ can be determined via forward and inverse kinematics. 
%All the possible types of robot to participate the task are represented by $C$, the actor space. $a \in A~(A \subset \mathbb{R}^3)$ denotes a 3-dimensional position where the end-effector of a robot actor $c \in C$ will move to. 

At each time step $t$, our system observes the environment state $(s_t,~\hat{s_t},~d_t,~b)$ and outputs $a_t$ for the given agent $c$. The agent $c$ moves to the given end-effector position, $a_t$. Thus, it pulls the rope, which moves the rigid object attached to it toward the target. Hence, DeRi-Bot aims to generate a series of actions $\{a_0,~a_1,~\dots,~a_T\}$, from $t=0$ to $t=T$, for each agent such that they minimize the Euclidean distance, indicated as $l$, between the centroid of the object's current position and the given target position.
% The rop output action indicates the desired end-effector position, th moves to the en  before and after actions.% between two consecutive actions. \textcolor{red}{We define the a time point between two adjacent actions. The t$^{th}$ step is before the t$^{th}$ action and after the t-1$^{th}$ action. Thus, the depth and segmentation map at step t are defined as $d_t$ and $s_t$.}
%It is worth noting that because of the control limitation brought by the high degree of freedom of ropes, we only consider the object's position and not the orientation. %Furthermore, we only use a 2D top-view orthographic projection of the target object with arbitrary orientation (2D rotation), to align the input format of APN and CPN, which will be discussed in Section~\ref{subsec:apn} and Section~\ref{subsec:cpn}.
%Formally, the objective of our task is the following: In an operational space that can be fully observed by a top-down view sensor, an object and a target is initialized respectively with a random pose. At each step $t$, the current environment state is available as $d_t \in D$ and $s_t \in S$. DeRi-Bot aims to generate a series of actions $a_1,~a_2,~\dots,~a_n$ to minimize the euclidean distance $l \in L~(L \in \mathbb{R})$ between the centroid of the object and the target. 

\subsection{Action Prediction Network (APN)}
\label{subsec:apn}
The objective of APN for a given agent is to predict the next action that reduces the distance between the object's current and the target's position given the environment state. Specifically, at step $t$, the input for APN is a 3-channel 2D array (depth map $d_t$, segmentation map $s_t$ and binary target map $b$):
\[
a_t \gets \text{APN}(d_t,~s_t,~b)
\]
All three channels are spatially pixel-aligned, which has been proved to help the deep neural network interpret the spatial relation among channels encoding different information~\cite{wang2021spatial,wu2020spatial}. Note that the $b$ will not change over the runtime for a given target. Furthermore, these channels are rotated so that the corresponding robot always resides on the top side, which also applies to the inputs to our CPN network. The output of APN $a_t = (x,~y,~z)$ represents a position that the end-effector of the corresponding robot will move to, where $x$, $y$, and $z$ is in the frame defined by the robot's operational space controller. The network is supervised by the Mean Square Error (MSE) between the network output and the ground truth command. The ground truth commands are generated by our random exploration strategy defined in Section III-E.

The model architecture of APN employs a Convolutional Neural Network (CNN)~\cite{lecun2015deep} to encode and interpret the 2D inputs. The intermediate output from the CNN is hereby processed by a Multi-Layer Perceptron (MLP) to output the desired 3D vector. In this work, the APN is backboned by a ResNet-style skip connected CNN~\cite{he2016deep}. The MLP module is a one-hidden layer fully connected network. 

As we mentioned in Section~\ref{sec:intro}, DeRi-Bot improvises based on the output of APN to ameliorate the stochastic problems induced by the deformability of soft bodies to improve the chances of success in reaching the target. Therefore, we model by a multivariate Gaussian distribution $X \sim \mathcal{N}(\mu,\,\Sigma)$, where $\mu = a_t^0$ and $ \Sigma = \sigma^2 * I$. We choose $\sigma$ = 0.5 and generate various $m$ extra action samples, $\{a^1_t,~\cdots,~a^{m}_t\}$, from that distribution which are then evaluated based on CPN predictions to select the best action $a^*_t$ for execution. The $a^0_t$ action is set to be $a_t$ generated originally by the APN.  

\subsection{Configuration Prediction Network (CPN)}
\label{subsec:cpn}
CPN takes as input the depth map $d_t$, the segmentation map $s_t$ and the action command $a^i_t \in \{a^0_t,~a^1_t,~ \cdots,~a^{m}_t\}$ to predict the outcome environment state segmentation map $\hat{s}^i_{t+1}$ after executing the action $a^i_t$:
\[
\hat{s}^i_{t+1} \gets \text{CPN}(d_t,~s_t,~a^i_t)
\]
The third channel $a^i_t$ is produced by broadcasting and zero-padding $a^i_t$ to the same size as other channels. We employ the deeplabv3+~\cite{chen2018encoder}, an advanced image segmentation method, to serve as the network structure of CPN. The network is trained using the supervision of the Binary Cross Entropy (BCE) loss between the network output and the ground truth segmentation map without rope encoding.

Although DeRi-Bot can work without CPN by simply iteratively executing the output of APN of each robot, the system's performance can be improved by the synergy of APN and CPN. Furthermore, the advantage of CPN is twofold. First, it allows the selection of the best action $a^*_t$ from a given set of actions, $\{a^0_t,~a^1_t,~\cdots,~a^{m}_t\}$, for a given agent, i.e., 
\[
a^*_{t} \gets \arg \min_i\; l(\text{CPN}(d_t,~s_t,~a^i_t),~b)\;\; \forall i \in [0,~m]
\]
Second, it also allows the selection of an agent during multiagent operations, i.e., the agent whose best action results in a minimum distance of the object to the goal among all agents.

\subsection{Workflow}
\label{subsec:flow}
The workflow of DeRi-Bot is presented in Algorithm~\ref{algo} and illustrated in Fig.~\ref{fig:workflow}. For a given task, at each step, APN proposes an action command for each of the $K$ agents given $d$, $s$, and $b$. A Gaussian sampler is employed to sample user-defined $m$ extra commands for each proposed action, yielding total $K \times (m+1) $ commands. Then, the generated commands will be sent to CPN to predict their corresponding outcome configurations. Given the outputs, we pick the action predicted to yield the shortest $l$ and execute it with its corresponding robot. It worth noting that we restrict the max velocity of the end-effector to a low value so that there is almost no sliding/drifting, i.e. acceleration/deceleration, motion involved. At the end of each step, the system can repeat the process or terminate if goal conditions are met, as described in Section~\ref{sec:exp}-B. However, in the case of a human-robot collaboration setting, the human participant gets the preference to decide if they want to act or let the robot executes their action. Furthermore, humans can also terminate the iterative process if they consider goal conditions have been met or the maximum number of interactions steps are reached without achieving the goal.% have been performed bu the  also if present in the system, are requested to use their own intelligence and judgement to perform these tasks. Human actors have the top priority, which allow them to access commands and corresponding predicted outcome for all robots, and to terminate a trial.

\vspace{-3mm}
\begin{algorithm}[h]
\caption{DeRi-Bot Framework}
\begin{algorithmic}
% \Require $n \geq 2$
% \Ensure $y = x^n$
\State Init(Object, Target)                              \Comment{Randomly place object and target}
\For{$ t = 0~\dots~\infty $}
\For{j = (1 \dots~k)}                                    \Comment{For each of the k robots}
\State $d_t,~s_t,~b \gets \text{Observe}(c_j)$      \Comment{Get Env config}
\State $a_t^0 \gets \text{APN}(d_t,~s_t,~b)$
\State $\hat{s}^{0}_{t+1} \gets \text{CPN}(d_t,~s_t,~a_t^0)$
\State $l_0^j \gets \text{Dist}(\hat{s}^{0}_{t+1},~b)$
    \For{i = (1 \dots~m)}                                \Comment{Sample m extra actions}
        \State $a_t^i \gets \text{Sample}(a_t^0,~\sigma)$ 
        \State $\hat{s}^{i}_{t+1} \gets \text{CPN}(d_t,~s_t,~a_t^i)$ \Comment{Predict outcome}
        \State $l_i^j \gets \text{Dist}(\hat{s}^{i}_{t+1},~b)$ \Comment{Calculate offset}
    \EndFor
\EndFor
\State $\hat{i},~\hat{j} \gets \text{argmin}(l_0^1~\dots~l_{m}^k)$  \Comment{Find best action index}
\State $\text{Execute}(a^{\hat{i}}_t, ~c_{\hat{j}})$        \Comment{Execute the best action}
\If {Term. Cond.} {break} \Comment{Break condition check}
\EndIf
\EndFor
\end{algorithmic}
\label{algo}
\end{algorithm}
\vspace{-5mm}

\begin{figure*}[t]
  \centering
  \vspace{10pt}
  \includegraphics[width=\textwidth]{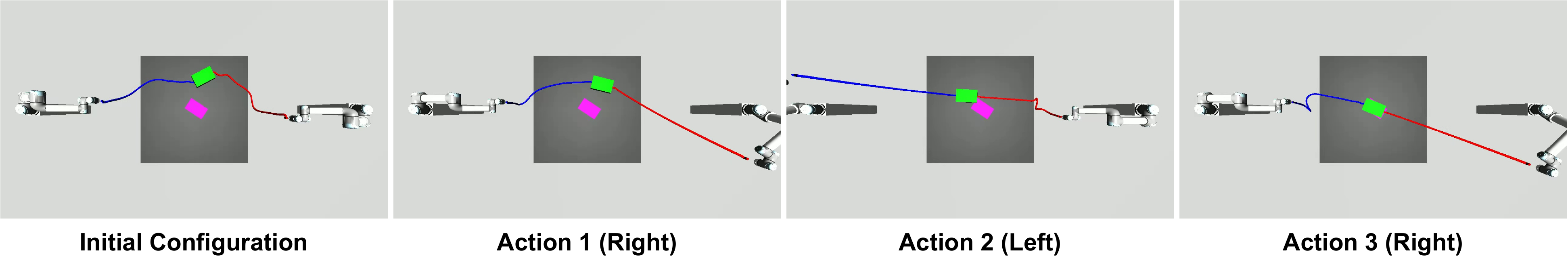}
  \caption{An example of a dual-bot setup in moving the green box to its target position indicated by the purple marker.}
  \label{fig:dual-exp}
  \vspace{-3mm}
\end{figure*}

\begin{table*}[t]
\caption{Evaluation results on a dual-robot simulation setup}
    \begin{center}
    \vspace{-3mm}
    \resizebox{0.80\textwidth}{!}{
        \begin{tabular}{l | c | c c c }
        \toprule
                        & \textbf{DeRi-Bot (APN + CPN)}        & Random Sampling    & Informed Sampling   & APN without CPN       \\ \midrule
        Shortest Offset & \textbf{0.306} $\pm$ \textbf{0.198}  & 0.393 $\pm$ 0.196  & 0.416 $\pm$ 0.228   & 0.317 $\pm$ 0.188 \\
        Final Offset    & \textbf{0.315} $\pm$ \textbf{0.207}  & 0.654 $\pm$ 0.268  & 0.540 $\pm$ 0.248   & 0.422 $\pm$ 0.235 \\
        Step Cost       & 5.29 $\pm$ 1.75    & 4.53 $\pm$ 1.11    & \textbf{4.35} $\pm$ \textbf{1.228}    & 4.70 $\pm$ 1.14   \\ \bottomrule
        \end{tabular}}
     \end{center}
     \label{tb1}
    \vspace{-6mm}
\end{table*}

\subsection{Data Generation}
\label{subsec:dg}
Our dataset was generated using a dual-bot setup, as illustrated in Fig.~\ref{fig:dual-exp}. In this setup, two robots take turns executing randomly sampled action commands. For each move, we randomly sample an $a$ within the valid space, and the robot moves its end-effector from its home position (shown as the top-left of Fig.~\ref{subfig:title_down}) to the sampled position, following the linear path between them. After each action, we collect depth maps before and after the movement and use color and depth information to produce corresponding segmentation maps. A dataset instance consists of the command, the depth, and the segmentation maps of the environment before and after the action, i.e., $d_t$, $s_t$, $d_{t+1}$, $s_{t+1}$, and $a_t$. Furthermore, in each instance of interaction, the box size (length, width, and height) was randomized within the allowed range for sim2real generalization. We choose box as our object type since we believe other shapes of objects can be abstracted by their corresponding bounding boxes. We expect that a large scale of interactive data points can help our neural networks interpret the underlying pattern governing the relationship between the environment configurations and actions.

To train the APN, as indicated in Section~\ref{subsec:apn}, the input data is $d_t$, $s_{t}$ and $b$ while the label is $a_t$. On the other hand, the CPN dataset instance has input data $d_t$, $s_t$ and $a_t$ with the ground truth of next state without rope information, i.e., $\hat{s}_{t+1}$. We collected 67,000 instances and divided them into training, validation, and testing splits with a ratio of 0.7 : 0.15 : 0.15.
%During the data collection process, we randomize the depth, width and height of the target object for generalization.

\begin{figure*}[t]
  \centering
  \vspace{10pt}
  \includegraphics[width=\textwidth]{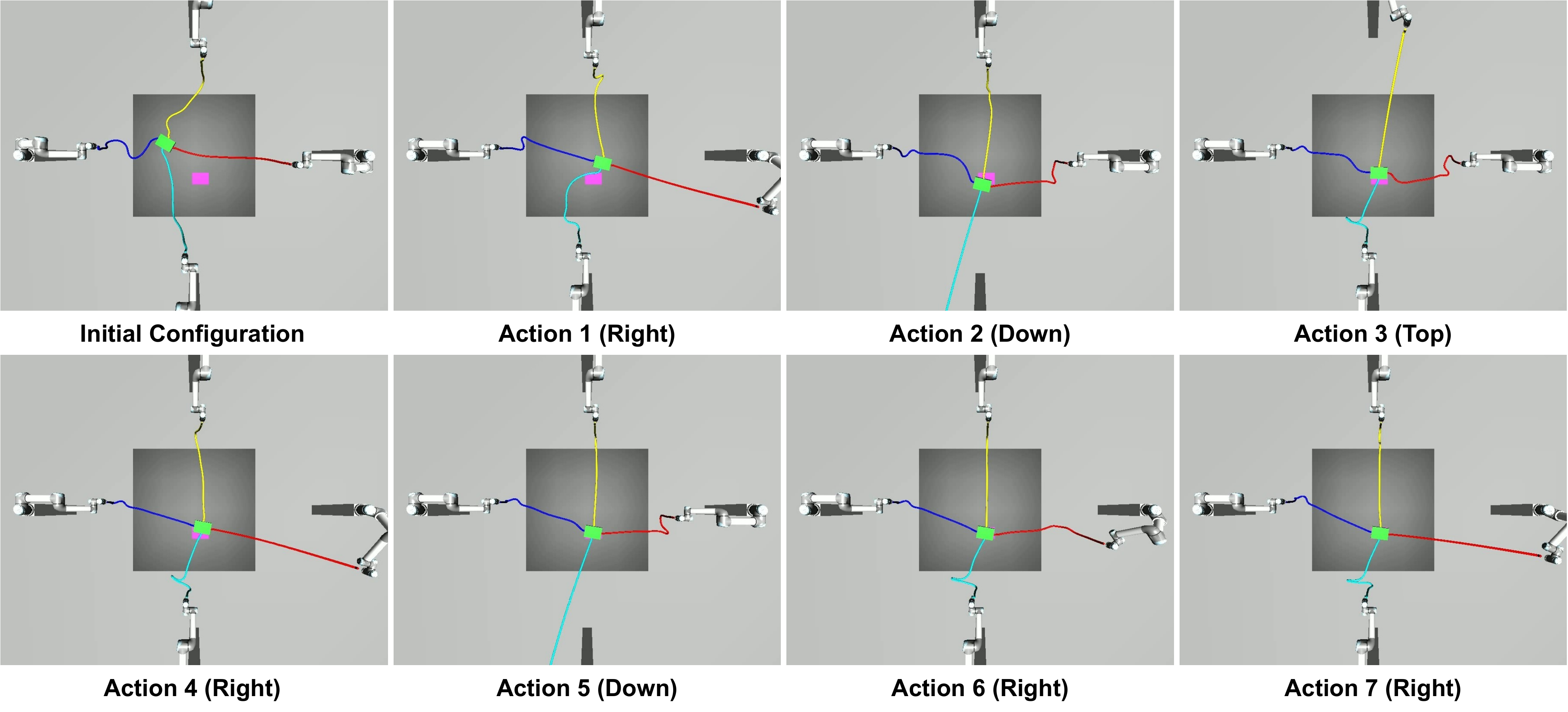}
  \caption{An example of the quad-bot collaboration process in moving the green box to its purple target. All robots propose their actions, and the robot yielding the minimum distance to the target is selected based on the CPN visual foresight.}
  \label{fig:quad-exp}
  \vspace{-2mm}
\end{figure*}

\section{Experiments}
\label{sec:exp}
We conducted a series of experiments in simulated environments for the dual-robot and the quad-robot setup and in the real world for the human-robot setup to evaluate the performance and generalization ability of our proposed DeRi-Bot framework. We comprehensively report the details of the setup and results of the experiments, along with the training and testing details of the individual neural models.

\subsection{Training and Evaluation of Neural Models}
\label{subsec:td}
We constructed and trained APN and CPN based on PyTorch-Lightning~\cite{falcon2019pytorch}, respectively, with a batch size of 100 using Adam Weight Decay (AdamW) optimizer~\cite{Loshchilov2017DecoupledWD}. The training process cost 1.53 and 0.73 hours for APN and CPN with a Nvidia RTX 3090 GPU. The final MSE loss of APN on the training and the test set is $0.0201$ and $0.0189$ meters. For the CPN, the final BCE loss on the training and the test set is $0.0060$ and $0.0061$, respectively. This validates that our neural model design does not overfit the training data and generalize to novel scenarios of the testing dataset. 

\subsection{Workflow Evaluation}
\label{subsec:ep}
We build a 2 meters $\times$ 2 meters square arena using MuJoCo~\cite{todorov2012mujoco}. In this area, create two test settings. First, the dual-robot setting contains two UR5e robots residing on the left and right sides of the arena. Second, the quad-robot setting contains four UR5e robots placed on each side of the arena. The object to be manipulated is connected to each robot by a rope. We create 100 test scenarios in each dual and quad robot setting. Each scenario contains a different-sized rigid object with its randomly sampled start and goal configurations. In these scenarios, we evaluate various baselines along with the DeRi-Bot. The terminal condition in each scenario is defined in terms of the shortest distance of the rigid object to its target. If more than three times the offset of the rigid object to the target position is greater than the history's shortest distance, the experiment is terminated, and corresponding evaluation metrics are recorded. 
%will be terminated. In our standard dual-robot setup and quad-robot generalization experiment, each method is asked to play 100 trials of such task.}

\subsection{Baselines}
\label{subsec:bl}
We evaluate our method along with other baselines to measure the relative performance of DeRi-Bot. The following methods are evaluated and presented in our sim-world experiments:
\begin{itemize}[leftmargin=*]
    \item \textbf{Random Sampling}: This method uniformly samples actions from the action space.
    \item \textbf{Informed Sampling}: This method operates as follows. Initially, we convert the vector, pointing from the object's current to the target position, from Cartesian ($x$, $y$, $z$) to polar ($\theta$, $r$) frame. We then divide the range of $2\pi$ radians into 12 equal bins defined by the lower and upper bound angles ($l_{1\dots12}$ and $u_{1\dots12}$). Next, we sample a real number $\hat{\theta}$ from a normal distribution: $$\hat{\theta} \sim \mathcal{N}(\theta,\,0.5) ,\;\; l_i < \hat{\theta} < u_i,$$ where $i \in [1,12]$ represents the index of the bin into which $\theta$ falls. Additionally, we scale $r$ by a factor of 3 and define a normal distribution around it to sample $\hat{r}$. The $\hat{r}$ is restricted to be greater than zero and the scaling factor is picked via cross-validation based on our observation. Finally, we get the sampled action ($\hat{x}$, $\hat{y}$, $\hat{z}$) by converting $(\hat{\theta}, \hat{r})$ to Cartesian frame. The end-effector will move to ($\hat{x}$, $\hat{y}$, $\hat{z}$) defined in its corresponding robot end-effector frame.
    \item \textbf{APN without CPN}: This method alternately executes commands generated by APN of each robot without sampling and CPN.
    \item \textbf{DeRi-Bot}: Our proposed framework incorporating the APN, sampling from a Gaussian Distribution, and CPN's visual foresight. We sample 3 extra actions from Gaussian Distribution around the output action from APN. 
\end{itemize}

\begin{table}[t]
\caption{Generalization Results of Quad-robot Setup}
    \begin{center}
    \vspace{-3mm}
    %\resizebox{0.48\textwidth}{!}{
    \begin{tabular}{l | c c}
        \toprule
                        & DeRi-Bot (4 Movable)  & DeRi-Bot (2 Movable)       \\ \midrule
        Shortest Offset & \textbf{0.117 $\pm$ 0.083} & 0.330 $\pm$ 0.164     \\
        Final Offset    & \textbf{0.127 $\pm$ 0.103} & 0.347 $\pm$ 0.176     \\
        Step Cost       & 5.25 $\pm$ 1.53       & \textbf{4.52 $\pm$ 1.51}   \\ \bottomrule
        \end{tabular}%}
    \end{center}
    \label{tb2}
    \vspace{-7mm}
\end{table}

\subsection{Metrics}
\label{subsec:mt}
The following metrics are employed to measure the performance of DeRi-Bot and the baselines quantitatively. Furthermore, we compute the mean and standard deviation across all test scenarios of these metrics:
\begin{itemize}[leftmargin=*]
    \item \textbf{Shortest Offset}: This metric represents the shortest distance in the trajectory between the object position and the target configuration. The unit of this metric is also meters.
    \item \textbf{Final Offset}: This represents the distance between the centroid of the target configuration and the object position after termination. The unit of this metric is in meters.
    \item \textbf{Step Cost}: The total step cost, i.e., the number of actions executed in each experiment.
\end{itemize}

\begin{figure*}[t]
  \centering
  \vspace{10pt}
  \includegraphics[width=\textwidth]{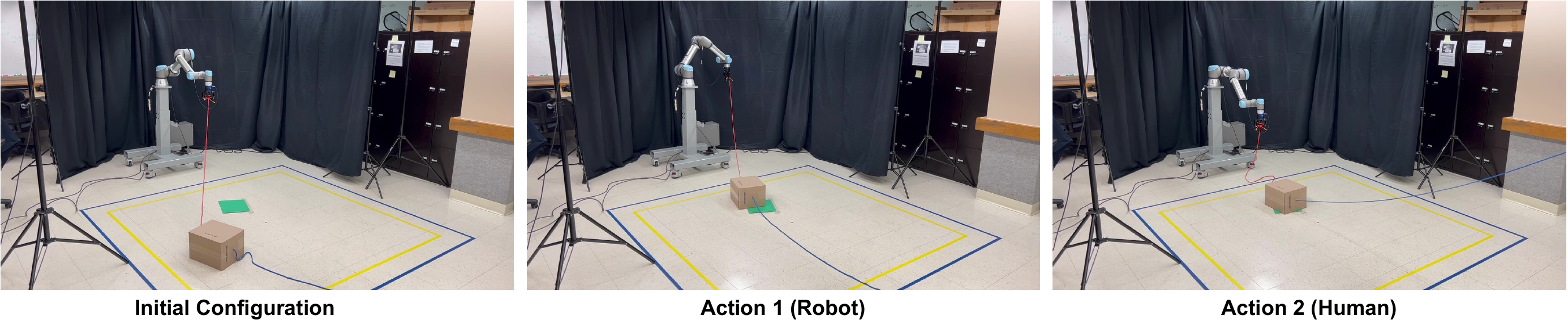}
  \caption{An example of short-horizon human-robot collaboration. It takes two steps to move the object to the target position.}
  \label{fig:real_exp_1}
\end{figure*}

\begin{figure*}[t]
  \centering
  \includegraphics[width=\textwidth]{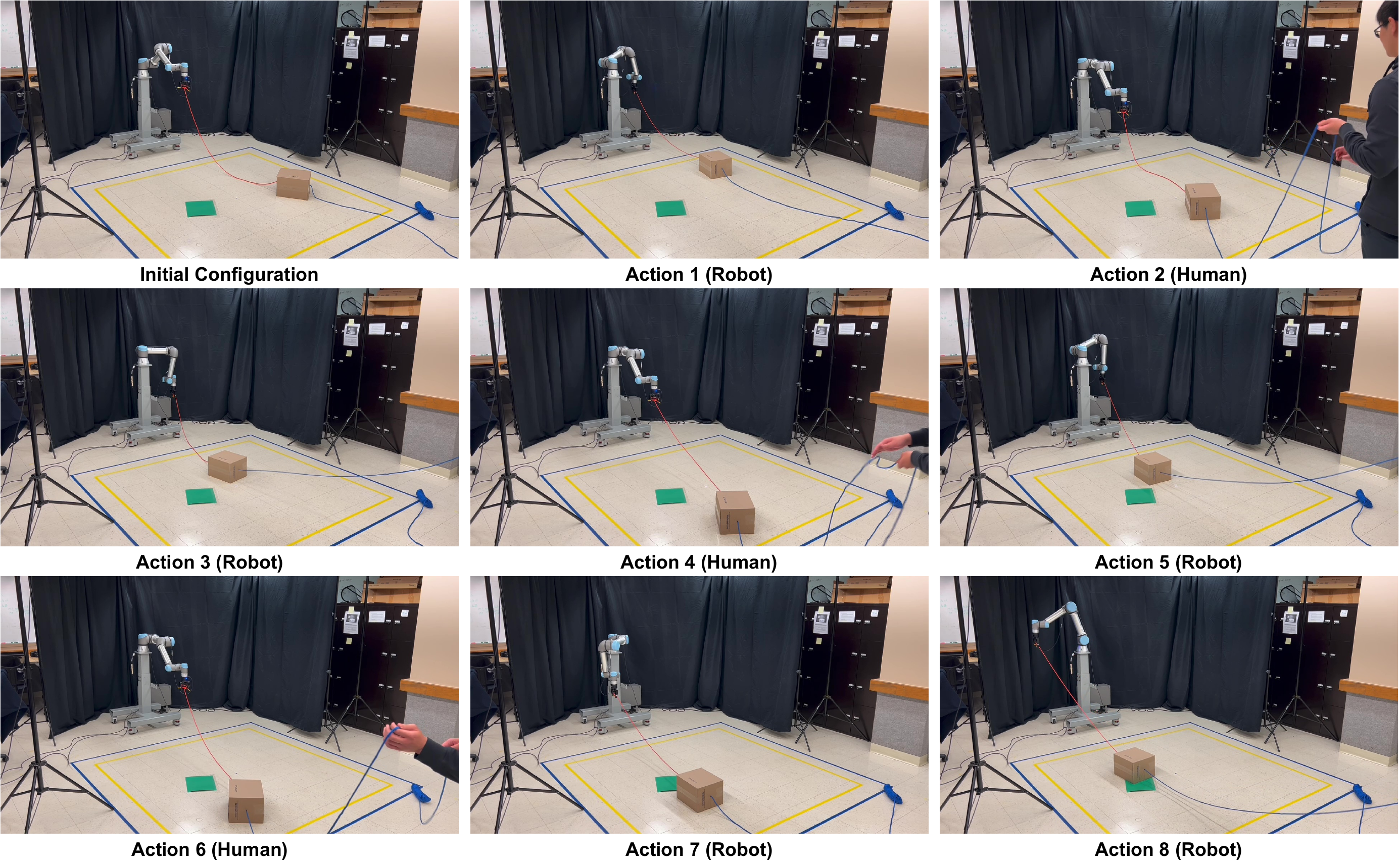}
  \caption{An example of long-horizon human-robot collaboration. It takes eight steps to move the object to the target position.}
  \label{fig:real_exp_2}
  \vspace{-2mm}
\end{figure*}

\subsection{Sim-world Results and Analysis}
\label{subsec:swr}

The experimental results of the standard dual-robot setup evaluation are presented in Table~\ref{tb1}, while an example of the task's top-down view is illustrated in Fig.~\ref{fig:dual-exp}. As Table~\ref{tb1} indicates, DeRi-Bot performs best in the experiments. Regarding the Shortest Offset metric, the APN-related methods, DeRi-Bot and APN w/o CPN, exhibit significant advantages over other baselines, indicating the robustness and effectiveness of the APN network. Furthermore, the better performance of DeRi-Bot compared to APN without CPN indicates that our sampling module with CPN foresight helps further reduce the offset and obtain the best performance among all methods. In terms of the Final Offset, the trend is the same as the Shortest Offset. However, the advantage of DeRi-Bot over other baselines is even larger. As we use a threshold-based termination strategy, we can see DeRi-Bot's framework is able to generate more viable commands to maintain the progress achieved, which strongly supports the feasibility and efficiency of our proposed framework pipeline. Finally, in terms of step cost, though DeRi-Bot consumed the most step budget, given its advantages in the Final and Shortest Offset, we believe they are necessary and tolerable costs to accomplish the given task. In summary, our experiments demonstrate that DeRi-Bot is a robust and effective framework pipeline for the dual-robot setup task. The results validate the capacity of the APN network and the usefulness of the CPN-supported sampling module in manipulating soft-rigid object systems.

\subsection{Sim-world Generalization Experiment}
\label{subsec:sge}
Our dataset and standard evaluation are conducted in the dual-robot setup as shown in Fig.~\ref{fig:dual-exp}. To evaluate the generalization ability of the DeRi-Bot, we built a quad-robot environment to perform the same task. In these settings, we also report dual robot performance by disabling the two robots and keeping the robots on the left and right sides of the arena.  %Notably, due to the physical limitation, we are unable to keep the initial object configurations the same as the dual-robot setup experiments. Thus, the result in Table~\ref{tb2} cannot be directly compared with Table~\ref{tb1}. To evaluate the effect of adding more robots, we conduct another experiment that has four robots but only two of them, which resides the same position as the dual-bot situation, are allowed to take actions.
According to the quantitative result, our conclusion is twofold. Firstly, DeRi-Bot can generalize to multiagent setups as expected without any retraining. 

% This is due to our decoupling strategy and spatially aligning and rotating the network's input maps such that the robot is always residing on the top side of them during prediction. %with up Migrating from dual-bot setup, DeRi-Bot works with quad-bot setup without any modification to the system. 
Secondly, Table~\ref{tb2} indicates the results are much better than the performance in a dual-robot setup. It aligns the intuition as more robots at different positions provide more moving directions. For example, Fig.~\ref{fig:quad-exp} shows vertical movements, which in the dual-robot case won't be possible, leading to more interaction steps to reach the target. % and will need  use a zigzag path in a dual-robot setup (e.g., Fig.~\ref{fig:dual-exp}), can be achieved by more direct path by robots resides on the top and bottom sides. 
It is also worth noting that in Fig.~\ref{fig:quad-exp}, action 6 does not affect the object's position, but action 7 moves it closer to the target. This is because the object's position after action 5 is almost perfect, so in action 6, the generated action yield no movement. However, in the next attempt, i.e., in action 7, CPN determined a sampled action could further minimize the offset, so the framework had the right robot to execute that command. This case strongly necessitates the CPN-supported sampling module to achieve the targets with minimum errors.

\subsection{Real-world Generalization Experiment}
\label{subsec:rge}
To test our proposed framework's sim-to-real transfer and human-robot setup generalization ability, we conducted a following real-world experiment. We created nine testing scenarios with different goal configurations and invited three volunteers to collaborate with our DeRi-Bot in moving the rigid object to given targets via ropes. %The instruction of the basic interaction rule of the system, each volunteer is tasked to act to achieve the same objective as the sim-world experiments for 3 trails.
Furthermore, the human and robot iteratively took steps to move the object. The human participant could do one of the following actions whenever the robot finished its action. Manipulate the rope to move the object, skip and let the robot conduct the manipulation again, or terminate the experiment. It is worth noting that, unlike stationary robots, human operators can walk to the desired position and pull the rope, yielding larger action space.
 
The average final and shortest offset of all the experiments are 0.261 and 0.194 meters, and the average step cost is 6.67. We demonstrate two example trails in Fig.~\ref{subfig:title_down},~\ref{fig:real_exp_1} and~\ref{fig:real_exp_2} for short and long-horizon operations. We observed that our framework accomplished the task objective with human operators in most situations, as also evident from the results. This validates that our decoupling design of DeRi-Bot empowers it to collaborate with humans without any modification or neural model retraining. Furthermore, it also validates that our data generation strategy allows direct sim2real transfer. 
%Furthermore, we would like to highlight the following points. The behavior of the soft-rigid body system is difficult for humans to anticipate. During our experiment, we asked our volunteers to always manipulate the rope to move the object to target position as close as possible. However, sometimes the outcome is contrary to expectation, resulting in the termination of the trial before reaching the ideal position due to a limited number of attempts. Despite having a broader range of actions available to them, human operators still achieved worse results than the inter-robot outcomes. We believe that this reinforces the importance of our framework.

\section{Conclusions and Future work}
\label{sec:con}
In this work, we propose the DeRi-Bot framework for manipulating rigid objects using 1D soft objects for transportation. Our neural models and algorithmic pipeline achieve high performance in moving the objects to their target positions. Furthermore, the framework is adaptable to multi-agent settings and real-world human-robot collaboration tasks. In future work, we aim to tackle the manipulation of unknown rigid objects with orientation constraints. This would also require the system to reason about complex rigid object geometry and the high degree of freedom rope dynamics.

% We propose the DeRi-Bot framework for manipulating rigid objects using soft objects for transportation. Our neural models and algorithmic pipeline achieve high-performance object movement. Our framework is adaptable to multi-agent settings and real-world human-robot collaboration. Future work includes tackling manipulation of unknown rigid objects with orientation constraints, considering complex object geometry and rope dynamics.
%\section{ACKNOWLEDGEMENT}
%We would like to thank Hanwen Ren and Syed Talha Bukhari for valuable insights and efforts in enhancing the quality of this work.
\bibliographystyle{IEEEtran}
\bibliography{root}

\end{document}